\documentclass[runningheads]{llncs}
\usepackage[T1]{fontenc}
\usepackage{graphicx}
\usepackage{bbding}
\usepackage{multirow}

\usepackage{eso-pic}

\begin{document}
\AddToShipoutPictureFG*{%
  \AtPageUpperLeft{%
    \raisebox{-1.5cm}{%
      \hspace{2cm}%
      \parbox{0.8\textwidth}{%
        \itshape
Preprint. This version has not undergone peer review.
      }%
    }%
  }%
}

\title{AAS-RAIL: Improving Information Extraction for Asset Administration Shells through Retrieval-Augmented In-Context Learning}
\titlerunning{AAS-RAIL for Information Extraction}

%
%
\author{Janek Groß\orcidID{0000-0002-6306-711X}\Envelope \and
Jens Heidrich\orcidID{0000-0001-6967-4722}}
\authorrunning{J. Groß and J. Heidrich}
%
\institute{University of Applied Sciences Mainz, Mainz, Rhineland-Palatinate, Germany\\
\email{\{janek.gross,jens.heidrich\}@hs-mainz.de}} 
\maketitle              
\begin{abstract}
The Asset Administration Shell (AAS) is a cornerstone of Industry 4.0 and the Digital Product Passport, providing standardized digital representations of industrial assets. While manufacturers already maintain extensive technical product documentation, generating AAS instances from existing product datasheets remains a labor-intensive task because technical information is extracted from heterogeneous document structures and often involves company-specific terminology and conventions.

In this work, we present AAS-RAIL, a retrieval-augmented information extraction (IE) approach that automatically generates Asset Administration Shells from PDF product datasheets using large language models (LLMs). Instead of relying on a fixed set of few-shot examples, the proposed retrieval-augmented in-context learning (RAIL) approach retrieves LLM-generated extraction helpers from similar Asset Administration Shells to provide instance-specific in-context learning (ICL). This enables the model to adapt its extraction behavior to company-specific naming conventions and formatting styles without fine-tuning. Our core contribution is the dynamic selection of company-specific AAS examples for each datasheet, replacing static prompting with an extraction pipeline that adapts to instances and combines semantic retrieval and structured information extraction.

The proposed approach is evaluated on a collection of industrial product datasheets using a selection of open- and closed-weight LLMs. Experimental results show that RAIL consistently improves extraction quality over conventional few-shot prompting, yielding relative improvements of 30.4--52.4\%. These results demonstrate that our approach provides an effective  improvement for company-specific AAS generation.

\end{abstract}
\section{Introduction}
\begin{figure}
\includegraphics[width=\textwidth]{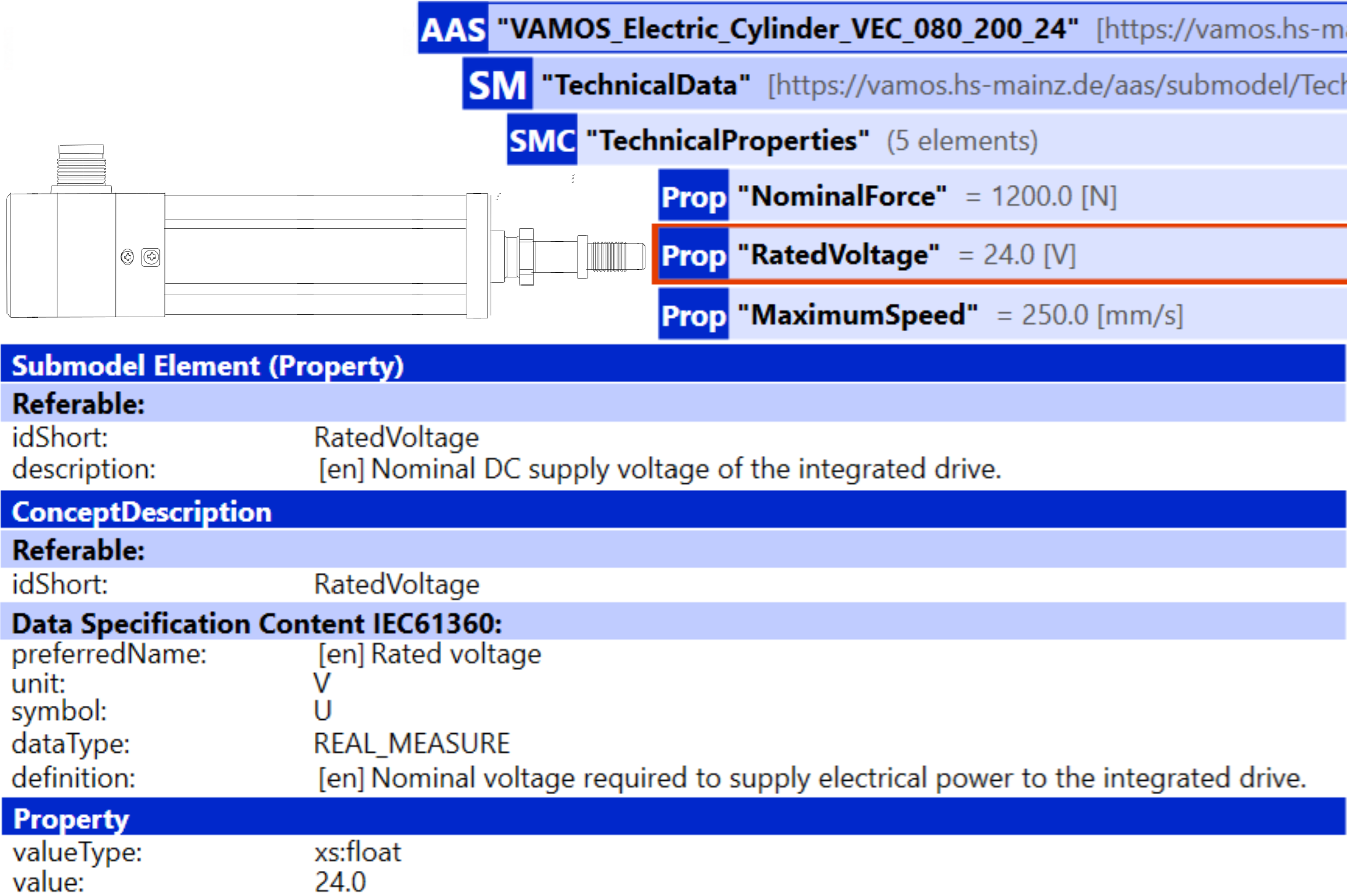}
\caption{Illustrative example AAS with \textit{TechnicalData} submodel, three technical properties and property metadata.} 
\label{fig1}
\end{figure}

Asset Administration Shells (AAS) \cite{IndustrialDigitalTwinAssociation.2023,IndustrialDigitalTwinAssociation.2023b} provide standardized digital representations of hardware and software assets (see Fig.~\ref{fig1}). By defining common semantics for properties, services, and metadata, they enable interoperable digital twins across heterogeneous Industry 4.0 environments. Beyond manufacturing, the AAS is increasingly considered a foundation for the Digital Product Passport, an initiative of the European Union to enable the exchange of product information throughout the entire product lifecycle. As the number and diversity of industrial assets continue to grow, scalable methods for creating and maintaining Asset Administration Shells become increasingly important.

An AAS consists of one or more submodels that refer to different aspects of an asset, like properties, documentation, or lifecycle data.
In this work, we focus on the \textit{TechnicalData} submodel, which contains collections of technical properties (e.g., rated voltage, operating temperature, or protection class). Each property is usually linked to a \textit{ConceptDescription} containing a property definition, data type, engineering unit, and additional metadata. AAS can be serialized in XML, JSON, or RDF (Turtle) format and are often distributed as AASX archives that contain the model and supplementary files.

While manufacturers can often generate AAS directly from internal engineering databases, this is typically not possible for legacy products or assets obtained from third parties. In these cases, the required information must be extracted manually from technical product documentation. This information extraction (IE) task aims to transform heterogeneous, semi-structured datasheets into machine-readable AAS representations. Recent advances in large language models (LLMs) have significantly simplified this process by enabling instruction-based extraction without task-specific training, resulting in several LLM-based AAS generation approaches \cite{Xia.2024,Vogel.2025,Kaya.2025}. However, technical product datasheets exhibit substantial variation in document structure, terminology, and company-specific conventions, making reliable extraction challenging. They may also contain data quality issues encountered in real-world applications. Language models may also generate plausible but unsupported property values or schema elements when the required information is missing, ambiguous, or difficult to align with the target AAS structure.

Manufacturers frequently model the same technical information using different property names or schema conventions while remaining compliant with the AAS specification. For example, one manufacturer may represent a product identifier as \textit{ArticleNumber}, whereas another uses \textit{OrderCode}. Static prompting struggles to capture such company-specific conventions reliably, whereas fine-tuning requires curated training data and repeated retraining as modeling conventions evolve.

In this work, we propose a retrieval-augmented IE approach that retrieves property-specific extraction helpers generated offline from previously modeled Asset Administration Shells. Each extraction helper summarizes how a particular property can be identified and transformed from a product datasheet. During extraction, helpers corresponding to semantically similar products are incorporated into the prompt as instance-specific in-context guidance. By conditioning the LLM on previously modeled assets with comparable characteristics, the proposed approach enables efficient, company-specific adaptation without modifying the underlying model parameters. The retrieved extraction helpers provide guidance on property identification, schema alignment, naming conventions, and output formatting while preserving the flexibility of prompt-based extraction.

This work addresses the following research questions:\\
\textbf{RQ1:} To what extent does retrieval-augmented in-context learning (RAIL) improve the quality of LLM-based information extraction for Asset Administration Shell generation?\\
\textbf{RQ2:} How do the number of retrieved examples and the resulting prompt context influence extraction performance and practical applicability?

The contributions of this paper are threefold. First, we propose a RAIL approach for company-specific AAS information extraction that leverages semantically similar examples and LLM-generated extraction helpers. Second, we present a comprehensive evaluation across multiple state-of-the-art open- and closed-weight LLMs, including an ablation study of the proposed retrieval components. The results demonstrate relative extraction improvements of approximately 30.4--52.4\% over conventional prompting strategies.
Third, we demonstrate that retrieval-augmented prompting can be applied without introducing substantial practical drawbacks by analyzing prompt context effects and changes in extraction outcomes.

The remainder of this paper is structured as follows. Section 2 summarizes the relevant background and related work. Section 3 describes the study design, the proposed approach and the experimental setup. Section 4 reports the experimental results, which are discussed in Section 5, Section 6 addresses threats to validity, and Section 7 concludes the paper.

\section{Related Work}
This section first summarizes the evolution from conventional to LLM-based information extraction before reviewing recent work on retrieval-augmentation for in-context learning and AAS generation.

Conventional information extraction consists of a series of standardized sub-tasks such as named entity recognition \cite{Seow2025}, relation extraction \cite{Zhao.2024}, and template or slot filling \cite{rombach.2025}. Classical systems often combine models that are trained for each specific task, domain-specific rules, and post-processing components to transform unstructured text into a predefined representation. Although such pipelines achieve high accuracy in well-defined domains, adapting them to new document types, entities, or output schemas usually requires corresponding labeled data, manual feature engineering, and retraining of models.
The use of LLMs for information extraction promises more flexibility to overcome these limitations.

The emergence of instruction tuning \cite{10.1145/3777411} for LLMs has enabled an end-to-end, generative approach to information extraction. While it was previously necessary to implement and train a separate component for each extraction sub-task, an LLM can be instructed to identify relevant entities, infer their relationships, and directly generate a target representation in a single inference process. Recent surveys describe this development as a transition towards generative information extraction \cite{Xu.2024GenerativeIESurvey,pai-etal-2024-survey}. This substantially reduces the effort required to prototype extraction systems for new domains, although empirical studies show that general-purpose LLMs often lag behind specialized state-of-the-art IE models \cite{han2023empirical} and remain sensitive to prompting, output constraints, and document complexity.

Instruction-tuned LLMs simplify information extraction, but the produced natural language text does not necessarily conform to predefined machine-readable schemas that are required for many IE tasks like the extraction of technical properties to create AAS. This requirement is addressed by grammar-constrained decoding \cite{geng-etal-2023-grammar} which restricts token generation to a context-free grammar. It forces the LLM to reliably produce structured outputs that follow a desired syntax such as JSON schemas without the need for rule-based post-processing. This capability makes grammar-constrained decoding a crucial element of generative information extraction pipelines.

While the correct output format can be enforced, extraction quality still depends on how the task is presented to the language model.
An important capability of modern LLMs is in-context learning (ICL), which allows them to adapt to new tasks based on examples included in the prompt instead of requiring parameter updates like during training or fine-tuning \cite{brown2020language,dong-etal-2024-survey}. This prompting approach is often categorized into zero-shot, one-shot, or few-shot prompting depending on the number of demonstrations in the prompt. While zero-shot prompting relies only on natural language instructions, one-shot and few-shot prompting additionally provide one or several input-output pairs that often allow the model to apply task-specific patterns directly from the prompt context. In a wide range of natural language processing tasks, few-shot prompting has consistently been shown to improve performance compared with zero-shot prompting.

The effectiveness of few-shot prompting depends on the selection of the demonstrations. Retrieval-augmented generation has become one of the most effective approaches for adapting LLMs to domain-specific tasks without retraining. In this approach, the prompt context is not fixed for all queries but parts of it are dynamically retrieved from a database. A similarity metric between the current input and the database entries is used to select the most relevant database entries to augment the prompt. Embedding-based retrieval has proven particularly effective because it provides task-relevant examples that are semantically similar to the current input instance. Recent surveys also identify retrieval-based demonstration selection as one of the most effective strategies for improving ICL while avoiding the cost of model fine-tuning \cite{luo2024context,dong-etal-2024-survey}.

While the previous work focuses on general IE techniques, several recent studies have applied instruction-tuned LLMs to the automated generation of AAS instances from technical documentation. Xia et al. generate AAS instances using an LLM-based multi-agent system and compare different prompting strategies for semantic model generation \cite{Xia.2024}. Kaya et al. and Vogel et al. likewise demonstrate that LLMs substantially reduce the manual effort required to create AAS-compliant representations from engineering documentation \cite{Kaya.2025,Vogel.2025}. While these approaches establish the feasibility of LLM-based AAS generation, they primarily rely on static prompting and generally target standardized AAS representations. As a result, they provide only limited support for adapting extraction to company-specific naming conventions or property schemas.

This work addresses these limitations by combining RAIL with LLM-based AAS information extraction. Instead of relying on a fixed set of demonstrations, the proposed approach retrieves properties from semantically similar AAS instances and incorporates them as extraction helpers for in-context learning. This enables the LLM to adapt to organization-specific schemas, naming conventions, and modeling practices without requiring model fine-tuning, while preserving the flexibility of prompt-based IE.

\section{Study Design}

The proposed approach\footnote{The implementation and links to the data are available at github.com/janek-gross/aas-rail} extends a conventional LLM-based AAS extraction pipeline by augmenting the extraction prompt with dynamically retrieved ICL examples. The retrieval database is constructed offline from existing AAS, while the extraction process itself is performed online.

\begin{figure}
\includegraphics[width=\textwidth]{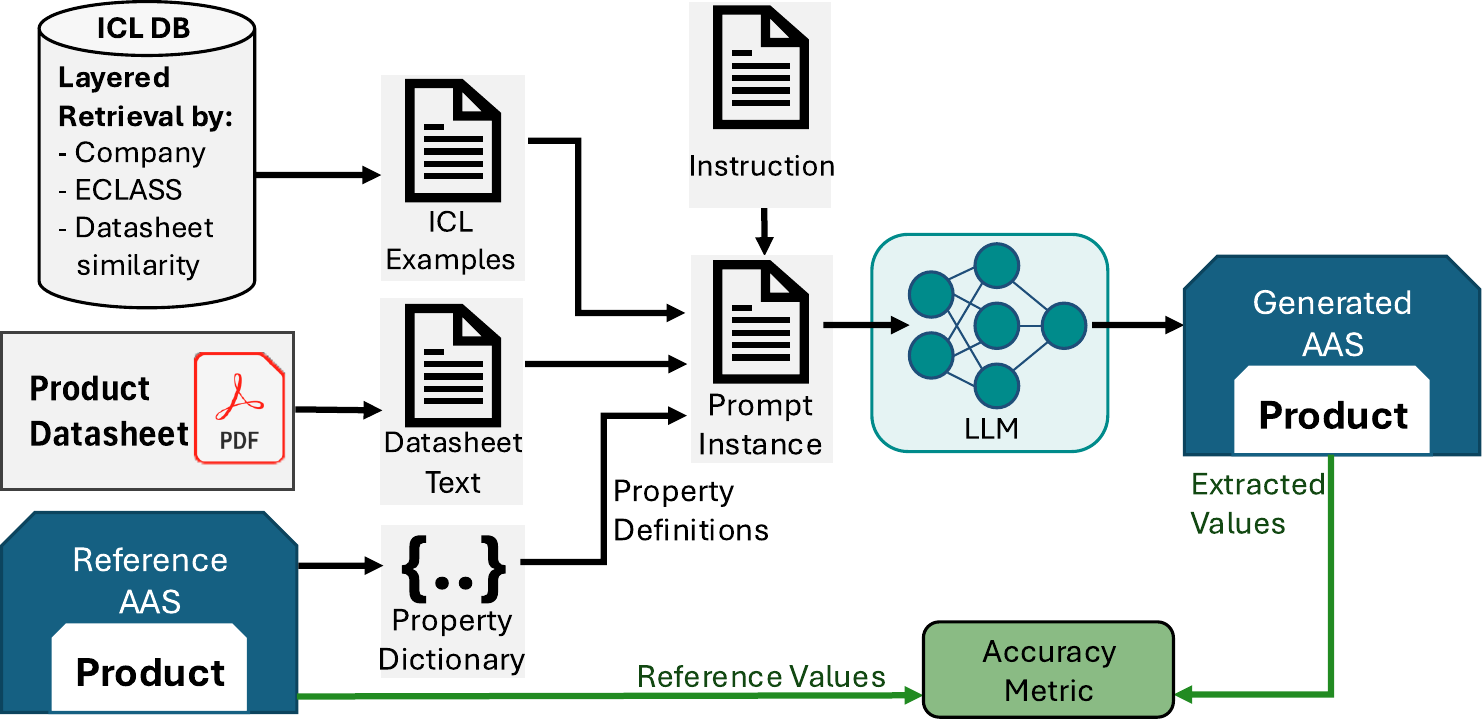}
\caption{Online inference pipeline for retrieval-augmented AAS extraction.} 
\label{fig2}
\end{figure}

Figure~\ref{fig2} illustrates the online IE pipeline. The product datasheet is first converted into textual input, while the corresponding reference AAS provides the property definitions that specify which values must be extracted. For each property, the layered retrieval mechanism selects relevant extraction helpers from the previously constructed ICL database based on manufacturer, ECLASS taxonomy, and datasheet similarity. These helpers are combined with the datasheet text, property definitions, and extraction instructions to form the prompt for the LLM, which generates the corresponding AAS values. Finally, the extracted values are compared with the reference values from the reference AAS to compute the accuracy metric.

\subsection{ICL Database Creation}
\begin{figure}
\includegraphics[width=\textwidth]{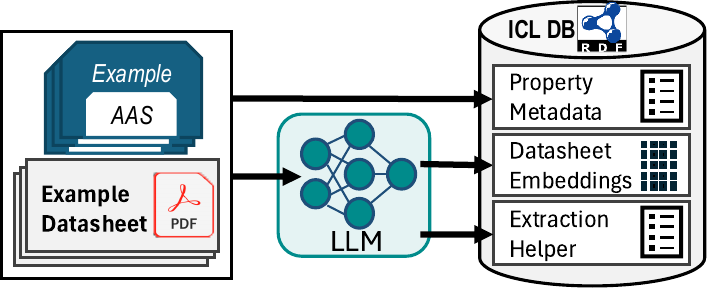}
\caption{Offline preparation of the ICL database.} 
\label{fig3}
\end{figure}

The offline phase of the proposed framework is illustrated in Figure~\ref{fig3}. Existing AAS instances are transformed into a retrieval database that stores property metadata, datasheet embeddings, and LLM-generated extraction helpers for later retrieval during online inference.

\subsubsection{Knowledge Database}
Existing AAS instances are provided as AASX archives containing XML or JSON representations of the Asset Administration Shell together with supplementary files. Since efficient retrieval requires structured access to individual properties and their metadata, all AASX files are converted into RDF format to create a knowledge graph.
This graph represents every technical property as an individual resource together with its associated metadata, including manufacturer information, ECLASS classification, ConceptDescriptions, and links to the embedding of the original product datasheet. The knowledge representation supports efficient retrieval of candidate properties during inference. This transformation is performed only once when the retrieval database is constructed.
\subsubsection{Generation of Extraction Helpers}
For every technical property contained in the RDF graph, an LLM generates an extraction helper describing how the property can be identified and extracted from its corresponding product datasheet. Unlike conventional few-shot examples, these helpers encode extraction guidance rather than merely providing an input-output pair.

\begin{table}
    \caption{LLM-generated extraction helper that describes how to extract the property named \textit{"Designation 1"} with value \textit{"MCB"} from the evidence text section. }
    \centering
    \begin{tabular}{p{0.35\textwidth}p{0.64\textwidth}}
        Extraction Helper Field & LLM generated Example \\
        \hline
        Grounding & exact \\
        Evidence & Header: "MCB Miniature circuit breaker". \\
        Extraction Rule & Use the abbreviated first term in the product header, preceding the spelled-out product type. \\
        Formatting Rule & Copy the abbreviation exactly, preserving uppercase. \\
        Avoid & Do not use the product code or article number as the designation.
    \end{tabular}
    \label{tab1}
\end{table}

To generate a helper, the LLM is given reference properties and the product datasheet. It is instructed to produce five fields for each property. The \textbf{Grounding} type field characterizes the relationship between the reference value and the datasheet as \textit{exact}, \textit{inferred}, \textit{conflicting}, or \textit{reference-only}. The \textbf{Evidence} field contains the text span that supports the respective reference value. The \textbf{Extraction Rule} field is used to describe where the value can typically be found. The \textbf{Formatting Rule} specifies how the extracted value should be transformed to fit the reference value including unit conversions and rounding. Finally, the \textbf{Avoid} field lists counter-examples of similar information that could be confused with the target value and should not be extracted. An example extraction helper is shown in Table~\ref{tab1}. Helper generation is only performed once during database construction and therefore does not contribute to inference time.

\subsection{Layered Retrieval of Extraction Helpers}
During inference, the system retrieves one or more candidate extraction helpers for each target property. When multiple candidates are available, they are ranked using a layered retrieval strategy. Instead of relying solely on semantic nearest-neighbor search, this strategy first prioritizes company-specific conventions and then considers semantic similarity. This ordering reflects the observation that naming conventions tend to vary more across manufacturers than within them.

The retrieval process consists of three stages. \\
\textbf{Company:} Properties originating from the same company receive highest priority because they most closely reflect organization-specific naming conventions and schema design. \\
\textbf{Product Similarity:} If multiple candidate properties exist, they are ranked according to their similarity within the ECLASS product taxonomy. We incorporate the four taxonomy levels \textit{Product Segment}, \textit{Product Group}, \textit{Product Class}, and \textit{Product Subclass} indicated by the product's ECLASS ID. Properties sharing deeper taxonomy levels are assumed to exhibit more similar technical characteristics.\\
\textbf{Datasheet Similarity:} Remaining candidates are ranked using cosine similarity between embeddings of the associated product datasheets. Datasheet embeddings are obtained by averaging the embeddings of individual text chunks generated during the ICL database creation.

Finally, the top-k ranked extraction helpers are inserted into the prompt as instance-specific in-context guidance.

\subsection{Dataset and Preprocessing}



After describing the proposed retrieval approach, we now introduce the experimental setup used for evaluation. The evaluation dataset consists of 200 pairs of industrial product datasheets and corresponding AAS from four different companies.
The reference AAS provides the target property definitions that specify which values should be extracted and it provides the reference values used to evaluate the extraction results. During deployment, only the predefined property schema is required; the reference values are used exclusively for evaluation.

To maximize product diversity, products were selected to cover as many ECLASS product categories as possible. The sampling procedure iteratively selected one product from each available ECLASS category.
Products were sampled iteratively across ECLASS subclasses. In each iteration, one product was selected at random from every subclass for which unsampled products remained. The iterations continued until the target sample size was reached. The resulting dataset spans multiple product categories in \textit{electrical engineering} and \textit{fluid power} for production automation.

An additional 40 datasheet–AAS pairs (10 per manufacturer) were selected using the same procedure to construct the ICL database. These examples are disjoint from the evaluation dataset and are used exclusively for retrieval to prevent information leakage.

Document preprocessing follows the procedure described in Groß et al. \cite{gross2026quality}. PDF datasheets are converted to text, while the corresponding AASX files are transformed into property dictionaries containing the property names, definitions, value types, and engineering units required for extraction.

\subsection{Experimental Procedure}
The primary objective of the evaluation is to investigate retrieval-augmented information extraction under realistic deployment conditions.
We therefore evaluate a selection of recent closed-weight LLMs and open-weight LLMs. The open-weight models were restricted to models that could be executed at a throughput of at least 20 tokens per second on a single NVIDIA DGX Spark system using 8-bit quantization. This restriction reflects the practical importance of local deployment for confidential industrial product data.
Since reasoning mode substantially increases inference latency for batched extraction, open-weight models are executed with reasoning mode disabled.
For each experiment the model that performs the information extraction is also used to create the extraction helpers.

Each product datasheet is processed once with RAIL and once without RAIL for every evaluated LLM, leaving all other variables constant.
The baseline corresponds to existing extraction pipelines employing instruction-based prompting together with a fixed set of few-shot examples. The proposed approach uses the identical prompt and few-shot examples but additionally augments the prompt with a retrieved extraction helper for each property.
The prompt is augmented with one retrieved extraction helper per property, if available, representing a sparse retrieval setting with at most one helper per property.

Further experiments vary the number of retrieved extraction helpers and the number of extracted properties per batch. Consequently, we investigate how the increased context length due to extraction helpers and due to batch size influences extraction performance.

For the evaluation, extracted values are first matched against the reference values of existing AAS using semantic similarity. Following Groß et al. \cite{gross2026quality}, non-numeric values with a cosine similarity $\geq 0.88$ are considered equivalent, while numeric values are considered correct if they differ from the reference value by no more than 1\%. The cosine similarity threshold was adopted from the previous work, where it was optimized to maximize correlation with controlled perturbations. These relaxed matching criteria account for semantically equivalent formulations, slight formatting differences, and minor rounding variations found between product datasheets and AAS references that hardly affect the engineering interpretation of the extracted information.

The task is evaluated as a fixed-slot value prediction problem with a predefined set of target properties. Grammar-constrained decoding ensures that the model produces exactly one output value for each property, which is then classified as either correct or incorrect with respect to the reference AAS. We therefore report the proportion of correctly populated properties as property extraction accuracy. Precision, recall, and $F_1$-score are not reported because property detection is outside the scope of the evaluation.

\section{Results}
\subsection{In-Context Learning Results}
Table~\ref{tab2} summarizes the accuracy achieved by the evaluated LLMs with and without RAIL. Across all tested models, incorporating retrieved in-context examples increased property extraction accuracy compared with the baseline. Relative improvements ranged from 30.4\% to 52.4\%, with an overall relative improvement of 38.4\%.

\begin{table}
\caption{Information extraction results with and without retrieval-augmented in-context learning. Results are averaged over individual AAS instances so that each product and company contributes equally regardless of the number of properties.}
\begin{tabular}{p{0.4\textwidth}p{0.2\textwidth}p{0.2\textwidth}p{0.2\textwidth}}
\multirow{2}{*}{Model} & \multicolumn{2}{l}{Property Extraction Accuracy \%} & Relative \\
 & baseline & with RAIL & Improvement \% \\
 \hline
 \hline
\multicolumn{4}{l}{\textbf{Open-weight models}} \\
GPT-oss 20B & 44.2 & 67.4 & \textbf{52.4}\\
Gemma-4 26B & 49.8 & 69.8 & 40.2 \\
Qwen-3.6 35B & 50.2 & 73.8 & 47.0 \\
\hline
\multicolumn{4}{l}{\textbf{Closed-weight models}} \\
GPT-5.4 Nano & 41.6 & 58.0 & 39.2 \\
GPT-5.6 Luna & 49.6 & 71.3 & 43.8 \\
GPT-5.6 Terra & 56.2 & 73.3 & 30.4 \\
GPT-5.6 Sol & 57.3 & 76.8 & 34.0 \\
Claude-5 Sonnet & 57.7 & 75.6 & 31.1 \\
Gemini-3.5 Flash & \textbf{59.6} & \textbf{79.3} & 32.9 \\
\hline
Overall & 51.8 & 71.7 & 38.4
\end{tabular}
\label{tab2}
\end{table}

All evaluated models benefited from RAIL. The three open-weight models improved by 40.2--52.4\%, while the closed-weight models improved by 30.4\%--43.8\%. With RAIL enabled, the achieved accuracy ranged from 58.0\% to 79.3\% across the evaluated models.

Table~\ref{tab3} summarizes the per-property effect of RAIL. For each extracted property, the prediction obtained with RAIL was compared with the corresponding prediction without RAIL.

\begin{table}
    \caption{Per-property change in the success of extractions through the use of RAIL.}
    \centering
    \begin{tabular}{p{0.27\textwidth}r}
        Condition & Percentage \\
        \hline
         Unchanged correct & 52.2 \\
         Improved & 22.4 \\
         Unchanged incorrect & 23.5 \\
         Worsened & 2.0 \\
    \end{tabular}
    \label{tab3}
\end{table}

For 52.2\% of all properties, both approaches produced correct extractions, while 23.5\% remained incorrect regardless of whether RAIL was used. RAIL converted previously incorrect predictions into correct ones for 22.4\% of all properties. Conversely, 2.0\% of properties that were extracted correctly without RAIL became incorrect when RAIL was applied.

\subsection{Ablation Study}
To assess the contributions of the extraction helpers and the layered
retrieval strategy, we conducted an ablation study using the Qwen-3.6 35B model.
First, the LLM-generated extraction helpers were replaced by corresponding
example values, while retaining the proposed retrieval strategy. Second,
the extraction helpers were retained, but the layered retrieval was replaced by random selection among properties with the same name.

\begin{table}
\centering
\caption{Ablation results using Qwen-3.6 35B.}
\label{tab:ablation}
\begin{tabular}{lr}
Configuration & Accuracy (\%) \\
\hline
Fixed few-shot baseline & 50.2 \\
Example values only, layered retrieval & 62.6 \\
Extraction helpers, random retrieval & 70.1 \\
\textbf{Proposed approach} & \textbf{73.8} \\

\end{tabular}
\end{table}

As shown in Table~\ref{tab:ablation}, using values from different AAS as examples in the prompt improved accuracy from 50.2\% to 62.6\%. However, replacing the complete
extraction helpers with example values alone reduced accuracy by 11.2 percentage
points compared to the proposed approach. Random retrieval caused a
smaller reduction of 3.7 percentage points. These results indicate that both
the generated extraction guidance and the relevance of the retrieved examples
contribute to the overall improvement.
\subsection{Effect of Context Length}
\begin{figure}
\includegraphics[width=\textwidth]{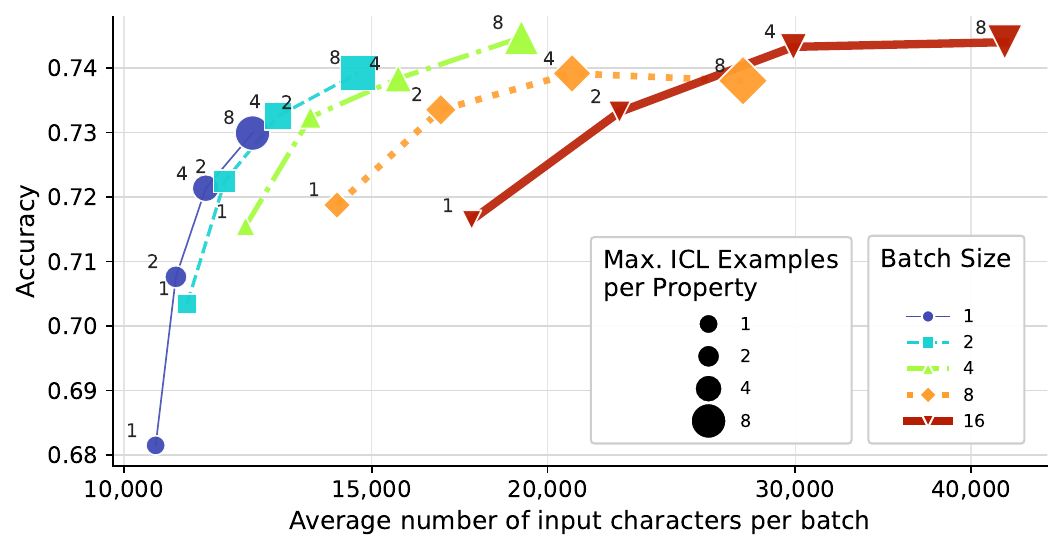}
\caption{Effect of increased context length depending on the number of simultaneously extracted properties and the number of extraction helpers per property.} 
\label{fig4}
\end{figure}

To investigate practical deployment considerations, additional experiments examined the influence of prompt context length on extraction performance. Figure~\ref{fig4} shows the achieved accuracy as a function of the number of simultaneously extracted properties (batch size) and the number of retrieved extraction helpers included in the prompt.

Across the evaluated configurations, accuracy generally increased as more retrieval examples were included. In contrast, the effect of batch size was less pronounced. The highest accuracy was obtained with four to eight retrieval examples per property. Differences among the larger batch-size configurations were comparatively small.

\section{Discussion}
The results provide a clear answer to RQ1. Across all evaluated LLMs, RAIL consistently improved extraction accuracy, demonstrating that dynamically retrieved extraction helpers provide effective guidance for company-specific information extraction. By conditioning the model on previously created AAS that reflect organization-specific terminology and conventions, RAIL enables adaptation to company-specific schemata without requiring model fine-tuning.

The ablation study further indicates that these improvements are not solely attributable to the presence of additional examples. Instead, a substantial part of the improvement originates from the richer LLM-generated extraction helpers, while the layered retrieval strategy provides an additional benefit by selecting more relevant demonstrations. Unlike conventional few-shot examples, the extraction helpers describe the extraction procedure rather than the expected output value, allowing the same helper to guide extraction for different products that share similar modeling conventions.

Although the best-performing configuration achieved an accuracy of 79.3\%, this is insufficient for fully automated AAS generation. However, this result should be interpreted in the context of the inherent limitations of the source documents rather than the extraction model alone. Manufacturers frequently omit technical properties, distribute relevant information across multiple document sections, or describe certain product details only implicitly. In an attempt to estimate how often the reference values were present in the evaluation dataset, we performed a regular-expression search that deliberately ignored ambiguities arising from multiple occurrences or unrelated matches. Despite these optimistic assumptions, the correct property value could be identified for only 56.2\% of all properties, while only 13.4\% contained both the property name and value within the document. Successful extraction therefore often requires at least some form of contextual reasoning rather than explicit key-value matching. As a consequence, the AAS-RAIL approach should be viewed as a supporting system to reduce manual modeling effort while leaving evaluation and correction to human domain experts.

RQ2 concerns the practical implications of RAIL, particularly the influence of retrieved examples and increasing prompt context length.
The additional experiments indicate that the increased prompt context introduced by RAIL does not substantially reduce extraction performance within the evaluated setting. Previous studies have shown that increasing the number of demonstrations can improve ICL performance \cite{brown2020language,luo2024context}, although more recent work suggests that longer demonstration sequences may fail to improve or even degrade performance for some tasks and models \cite{dong-etal-2024-survey}. Within the evaluated setting, however, no such degradation was observed. Instead, increasing the number of retrieved extraction helpers consistently improved extraction quality, indicating that the evaluated LLMs were able to effectively utilize the additional retrieved context. Although the optimal configuration still depended to some extent on the extraction batch size, these findings suggest that retrieval-augmented prompting can be scaled within practical context lengths without incurring significant drawbacks.

A further concern is whether retrieved demonstrations can bias the model towards incorrect predictions. Extraction failures caused by retrieval were rare: the use of RAIL corrected 22.4\% of previously incorrect extractions while only 2.0\% of previously correct predictions became incorrect. Overall, the improvement considerably exceeds the infrequent extraction failures.

Besides extraction accuracy, successful industrial deployment of LLM-based systems also depends on practical considerations such as maintainability, data confidentiality, solution architecture, and automated evaluation \cite{yu2025experience}. The proposed RAIL approach addresses some of these requirements by externalizing company-specific knowledge into a retrieval database instead of the model parameters. This allows new Asset Administration Shells to be incorporated incrementally without retraining while remaining compatible with locally deployed open-weight LLMs for proprietary engineering data. Consequently, the extraction system can evolve together with company-specific modeling practices without repeated fine-tuning of the underlying language model.

\section{Threats to Validity}

This study is subject to several threats to validity.

\textbf{Construct validity.}
Extraction quality is evaluated by comparing generated property values with reference AAS using semantic similarity. Although the embedding-based evaluation accounts for semantically equivalent representations, the selected similarity threshold may occasionally classify related but incorrect values as correct. The evaluation assumes that the reference AAS instances represent the intended ground truth. Ambiguous, incomplete, or inconsistent information contained in the original product datasheets may therefore also influence the reported extraction quality.
Furthermore, the reported results are point estimates without confidence intervals or statistical significance tests. Because predictions with and without RAIL are paired for the same properties, paired bootstrap confidence intervals or tests such as McNemar’s test could be used to quantify the uncertainty of the observed improvements.

\textbf{Internal validity.}
The observed improvements may not be attributable exclusively to the retrieval strategy. Retrieval-augmented prompts contain additional manufacturer-specific information and are generally longer than the baseline prompts. Although the separate context-length analysis indicates that increased prompt length alone does not explain the observed improvements, interactions between retrieval quality, prompt structure, and context length cannot be excluded completely. In addition, the retrieval database consists of 40 representative AAS examples, which may not fully reflect the size and diversity of retrieval collections encountered in industrial practice. Although the retrieval and evaluation sets contain disjoint datasheet–AAS pairs, products from the same manufacturer may belong to closely related product families or share similar datasheet templates. Consequently, near-duplicate documents or product variants could make retrieval easier and inflate the observed benefit.

\textbf{External validity.}
The evaluation focuses on the \textit{TechnicalData} submodel using technical product datasheets from four manufacturers. Consequently, the reported results may not directly generalize to other AAS submodels, additional document types, other industrial domains, or multilingual documentation. Although both open- and closed-weight LLMs were evaluated, the experiments cover only a limited selection of current models. Future model generations or substantially larger retrieval databases may exhibit different retrieval behavior and extraction characteristics. Furthermore, the evaluation assumes that examples are already available to create the retrieval database. The results therefore characterize a within-manufacturer setting and do not establish performance for new manufacturers for which no prior AAS instances exist.

\textbf{Conclusion validity.}
The reported conclusions are based on a single evaluation dataset and a fixed set of prompt configurations. While the consistent improvements observed across all evaluated LLMs increase confidence in the reported trends, additional studies using larger and more diverse datasets, alternative retrieval strategies, and repeated model executions would further strengthen the conclusions. In particular, evaluating the proposed approach across a broader range of manufacturers and engineering domains would provide stronger evidence for its general applicability.
Furthermore, closed-weight APIs may change model implementations over time even when model names remain stable. Consequently, exact reproduction requires recording model snapshots or version identifiers, API dates, and all available decoding parameters.

\section{Conclusion and Future Work}

This work presented a retrieval-augmented IE approach for generating AAS from technical product datasheets using LLMs. By dynamically retrieving guidance from relevant Asset Administration Shells and incorporating it into the prompting process, the proposed approach enables adaptation to company-specific naming conventions without requiring model fine-tuning.

The experimental evaluation demonstrated that RAIL improved extraction quality for every model in the evaluated configuration without major practical drawbacks. In addition, the results indicate that reasonably sized open-weight models achieve competitive performance, making local deployment of company-specific AAS generation systems a practical option for industrial applications.

Future work should investigate the integration of RAIL into agentic IE systems. Combining RAIL with iterative document exploration and self-evaluation represents a promising direction for further improving the reliability and completeness of automatically generated Asset Administration Shells.

\begin{credits}
\subsubsection{\ackname}
This work was supported by the research training group “Dependable AI Assistants for the Management of Dynamic Production Systems and Supply Chains (VAMoS)” at Mainz University of Applied Sciences and the Rheinland-Palatinate Technical University of Kaiserslautern-Landau, funded by the Ministry of Science and Health of Rhineland-Palatinate. The authors gratefully acknowledge this support.
\subsubsection{\discintname}
The authors have no competing interests to declare that are
relevant to the content of this article.
\end{credits}

\bibliographystyle{splncs04}
\bibliography{references}

\end{document}